\def\paperID{0000}
\def\confName{ICCV}
\def\confYear{2025}

\newif\ifreview 
\newif\ifarxiv \newcommand{\arxiv}{\arxivtrue}
\newif\ifcamera 
\newif\ifrebuttal 

\arxiv 

\pdfoutput=1
\documentclass[10pt,twocolumn,letterpaper]{article}
\ifreview \usepackage[review]{cvpr} \fi
\ifarxiv \usepackage[pagenumbers]{cvpr} \fi
\ifrebuttal \usepackage[rebuttal]{cvpr} \fi
\ifcamera \usepackage{cvpr} \fi

\usepackage{graphicx}	
\usepackage{amsmath}	
\usepackage{amssymb}	
\usepackage{booktabs}
\usepackage{times}
\usepackage{microtype}
\usepackage{epsfig}
\usepackage{caption}
\usepackage{float}
\usepackage{placeins}
\usepackage{color, colortbl}
\usepackage{stfloats}
\usepackage{enumitem}
\usepackage{tabularx}
\usepackage{xstring}
\usepackage{multirow}
\usepackage{xspace}
\usepackage{url}
\usepackage{subcaption}
\usepackage{xcolor}
\usepackage[hang,flushmargin]{footmisc}

\ifcamera \usepackage[accsupp]{axessibility} \fi

% ...

\ifarxiv  \fi

\newcommand{\R}[1]{{%
    \textbf{%
        \ifstrequal{#1}{1}{\textcolor{red}{R#1}}{%
        \ifstrequal{#1}{2}{\textcolor{blue}{R#1}}{%
        \ifstrequal{#1}{3}{\textcolor{magenta}{R#1}}{%
        \ifstrequal{#1}{4}{\textcolor{teal}{R#1}}{%
                           \textcolor{cyan}{R#1}%
        }}}}%
    }%
}}

\usepackage{xr-hyper}

\makeatletter
\newcommand*{\addFileDependency}[1]{
  \typeout{(#1)}
  \@addtofilelist{#1}
  \IfFileExists{#1}{}{\typeout{No file #1.}}
}

\makeatother

\definecolor{cvprblue}{rgb}{0.21,0.49,0.74}
\usepackage[pagebackref,breaklinks,colorlinks,allcolors=cvprblue]{hyperref}
\usepackage[capitalize]{cleveref}
\crefname{section}{Sec.}{Secs.}
\crefname{table}{Table}{Tables}
\crefname{figure}{Fig.}{Figs.}

\ifarxiv \crefname{appendix}{App.}{Apps.}
\else \crefname{appendix}{Suppl.}{Suppls.} \fi

\definecolor{cvprblue}{rgb}{0.21,0.49,0.74}
\usepackage[pagebackref,breaklinks,colorlinks,allcolors=cvprblue]{hyperref}
\usepackage{colortbl}
\usepackage{pifont}       
\usepackage{bbding}       
\usepackage{fontawesome}  
\usepackage{multirow}
\usepackage{multicol}

\def\paperID{7572} 
\def\confName{ICCV}
\def\confYear{2025}

\definecolor{Gray}{gray}{0.95}
\definecolor{barriercolor}{RGB}{255, 120, 50}
\definecolor{bicyclecolor}{RGB}{255, 192, 203}
\definecolor{buscolor}{RGB}{255, 255, 0}
\definecolor{carcolor}{RGB}{0, 150, 245}
\definecolor{constructcolor}{RGB}{0, 255, 255}
\definecolor{motorcolor}{RGB}{200, 180, 0}
\definecolor{pedestriancolor}{RGB}{255, 0, 0}
\definecolor{trafficcolor}{RGB}{255, 240, 150}
\definecolor{trailercolor}{RGB}{135, 60, 0}
\definecolor{truckcolor}{RGB}{160, 32, 240}
\definecolor{driveablecolor}{RGB}{255, 0, 255}
\definecolor{otherflatcolor}{RGB}{139, 137, 137}
\definecolor{sidewalkcolor}{RGB}{75, 0, 75}
\definecolor{terraincolor}{RGB}{150, 240, 80}
\definecolor{manmadecolor}{RGB}{213, 213, 213}
\definecolor{vegetationcolor}{RGB}{0, 175, 0}
\definecolor{otherscolor}{RGB}{0, 0, 0}
\definecolor{lightskyblue}{rgb}{0.53, 0.81, 0.98}
\definecolor{aliceblue}{rgb}{0.94, 0.97, 1.0}

\begin{document}
\title{$I^{2}$-World: Intra-Inter Tokenization for Efficient Dynamic 4D Scene Forecasting}
\author{Zhimin Liao, ~Ping Wei\textsuperscript{\rm}\thanks{Corresponding author.}, ~Ruijie Zhang, ~Shuaijia Chen, ~Haoxuan Wang, ~Ziyang Ren\\
National Key Laboratory of Human-Machine Hybrid Augmented Intelligence\\
Institute of Artificial Intelligence and Robotics, Xi’an Jiaotong University\\
{\tt\small liaozm@stu.xjtu.edu.cn,~pingwei@xjtu.edu.cn}}
\maketitle

\begin{abstract}
     Forecasting the evolution of 3D scenes and generating unseen scenarios via occupancy-based world models offer substantial potential for addressing corner cases in autonomous driving systems. While tokenization has revolutionized image and video generation, efficiently tokenizing complex 3D scenes remains a critical challenge for 3D world models. To address this issue, we propose $I^{2}$-World, an efficient framework for 4D occupancy forecasting. Our method decouples scene tokenization into intra-scene and inter-scene tokenizers. The intra-scene tokenizer employs a multi-scale residual quantization strategy to hierarchically compress 3D scenes while preserving spatial details. The inter-scene tokenizer residually aggregates temporal dependencies across timesteps. This dual design preserves the compactness of 3D tokenizers while retaining the dynamic expressiveness of 4D tokenizers. Unlike decoder-only GPT-style autoregressive models, $I^{2}$-World adopts an encoder-decoder architecture. The encoder aggregates spatial context from the current scene and predicts a transformation matrix to enable high-level control over scene generation. The decoder, conditioned on transformation matrix and historical tokens, ensures temporal consistency during generation. Experiments demonstrate that $I^{2}$-World achieves state-of-the-art performance, outperforming existing methods by \textbf{25.1\%} in mIoU and \textbf{36.9\%} in IoU for 4D occupancy forecasting while exhibiting exceptional computational efficiency. It nearly requires \textbf{2.9 GB} of training memory and achieves real-time inference at \textbf{37.0 FPS}. Our code is available on \url{https://github.com/lzzzzzm/II-World}.

\end{abstract}
\section{Introduction}
    \begin{figure}[t!]
    \centering
    \begin{subfigure}{0.47\textwidth}
        \centering
        \includegraphics[width=\linewidth]{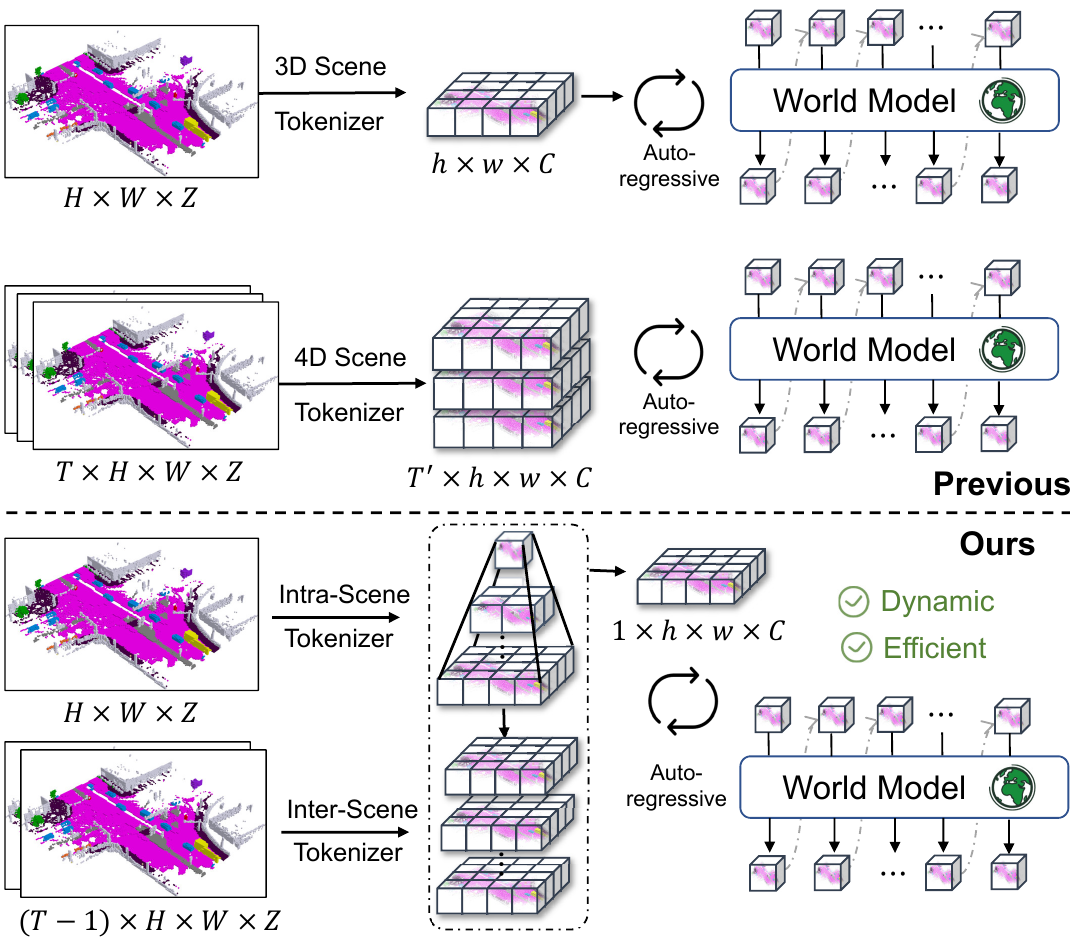}
        \caption{Our proposed tokenizer versus previous tokenizer methods.}
        \vspace{1em}
        \label{figure:introduction-tokenizer}
    \end{subfigure}
    \centering
    \begin{subfigure}{0.47\textwidth}
        \centering
        \includegraphics[width=\linewidth]{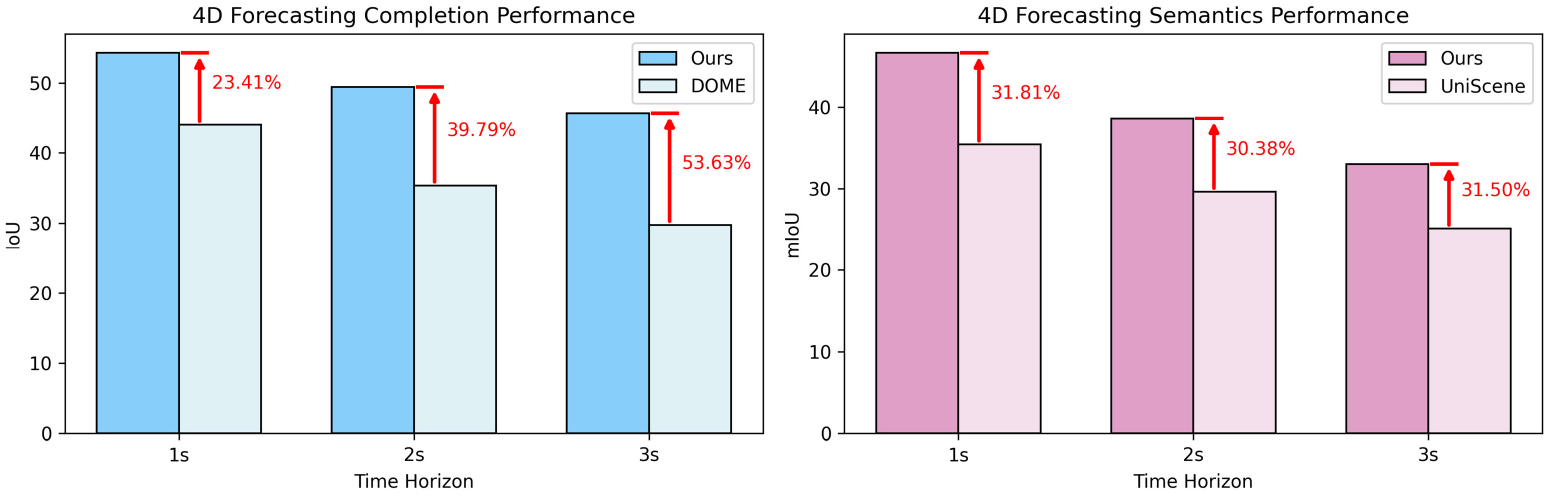}
        \caption{Comparison with the state of the art.}
        \label{figure:introduction-model-comparsion}
    \end{subfigure}
    \caption{\textbf{(a) Our proposed tokenizer versus previous tokenizer methods.} We decouple scene tokenization into intra-scene and inter-scene components, which enable our method to maintain the computational efficiency of 3D tokenizers while modeling dynamic information akin to 4D tokenizers. \textbf{(b) Comparison with the state of the art.} Our method significantly outperforms the previous state-of-the-art method, particularly in 3-second forecasting.} 
    \label{figure:motivation}
\end{figure}

    3D occupancy~\cite{voxformer, tpvformer, occ3d, semantickitti} offers more geometric and detailed information about 3D scenes, making it more suitable for autonomous driving systems than traditional 3D bounding boxes~\cite{bevdepth, bevdet, bevformer, bevformerv2, bevstereo, sparsebev} and point clouds~\cite{pointpillars, fpn, pointrcnn}. With the advancement of generative AI, occupancy-based world models~\cite{occworld, occ-llm, drive-occworld, driveworld, dome, occsora, vista} that generate and forecast 3D occupancy show great potential to serve as the world simulator and address corner cases~\cite{worldmodel-survey} in complex traffic scenarios.

    Inspired by recent advances in image and video tokenization~\cite{vq-vae,vq-vae-v2,cv-vae,vista,copilot4d}, the tokenization of 3D scenes has emerged as a foundational prerequisite for building 3D world models~\cite{occ-llm,occllama,occworld,dome}. Nevertheless, the efficient compression of 3D scenes into compact tokens while preserving spatiotemporal dynamics remains a critical challenge. As illustrated in~\cref{figure:introduction-tokenizer}, existing tokenizers can be broadly categorized into 3D scene tokenizers~\cite{occ-llm,occllama,occworld,renderworld} and 4D scene tokenizers~\cite{occsora,dome}. 3D scene tokenizers compress individual 3D scenes into highly compact latent representations. Although these tokenizers achieve accurate scene reconstruction, their inability to model temporal dynamics severely constrains their predictive capacity. By contrast, 4D scene tokenizers, inspired by video-generation frameworks~\cite{videogpt,cv-vae,copilot4d,vista}, directly process 4D spatiotemporal tokens, thereby embedding dynamic scene evolution within the token space. Although this approach enhances forecasting fidelity, the attendant high-dimensional tokens impose prohibitive computational overhead on downstream autoregressive models~\cite{gpt-4} and diffusion frameworks~\cite{stable-diffusion}, rendering them impractical for latency-critical applications such as autonomous driving.

    In this paper, we propose \textit{$I^{2}$-World}, an efficient world model designed for 4D occupancy forecasting. It consists of two principal components: the \textit{$I^{2}$-Scene Tokenizer}, which performs intra and inter frame scene tokenization and the \textit{$I^{2}$-Former}, an autoregressive framework guided by transformation matrixs.

    The \textit{$I^{2}$-Scene Tokenizer} decouples the tokenization process into two complementary components: an Intra-Scene Tokenizer and an Inter-Scene Tokenizer. The Intra-Scene Tokenizer adopts a multi-scale tokenization strategy~\cite{var,rq-vae}, focusing on capturing fine-grained details and static context within the current scene. In contrast, the Inter-Scene Tokenizer maintains a memory queue to store historical scene tokens and employs temporal quantization along the timestamp to model dynamic motion within the scene. By integrating these components, our $I^{2}$-Scene Tokenizer produces high-compression tokens comparable to those of 3D tokenizers while preserving the temporal modeling capabilities of 4D tokenizers.

    To incorporate the proposed \textit{$I^{2}$-Scene Tokenizer}, we introduce \textit{$I^{2}$-Former}, a hybrid architecture that differs from decoder-only GPT-like autoregressive models~\cite{gpt-4}. Instead, \textit{$I^{2}$-Former} consists of two core components: an Intra-Encoder and an Inter-Decoder. The Intra-Encoder hierarchically aggregates spatial context from the current scene tokens. In parallel, it leverages diverse ego-car actions to predict a transformation matrix that maps the current scene representation to the next timestep. The transformation matrix then serves as a conditional guidance signal for the Inter-Decoder, which dynamically integrates historical and current scene tokens to predict the next-timestep scene token.

    The decoupled design of $I^2$-Former enables fine-grained controllable generation. Users can intuitively steer scene predictions by manipulating the transformation matrices or by adjusting diverse control actions, thereby enabling flexible adaptation to a wide range of driving scenarios. Our contributions can be summarized as follows:
    \begin{itemize}

        \item We propose \textit{$I^{2}$-Scene Tokenizer}, which retains the computational efficiency of 3D tokenizers while achieving the temporal expressiveness of 4D tokenizers.  
        
        \item We introduce \textit{$I^{2}$-Former}, an autoregressive architecture conditioned on the transformation matrices that enables high-fidelity scene generation with enhanced details.

        \item Our \textit{$I^{2}$-World} advances 4D occupancy forecasting by \textbf{25.1\%} in mIoU and \textbf{42.9\%} in IoU relative to prior state-of-the-art approaches. Moreover, it achieves computational efficiency, requiring nearly \textbf{2.9 GB} of training memory and performing inference at \textbf{37.0 FPS}.
        
    \end{itemize}

\section{Related Work}
    \subsection{Occupancy-based World Model}
        World models~\cite{vista, copilot4d, magicdrive, drivedreamer, drivedreamer-2, differentiablepc, selfpc, proxypc, SPF2} leverage an agent’s actions and historical observations to forecast the evolution of future 3D scenes, enabling the agent to reason about dynamic environments. Recent advances in occupancy-based world models have explored diverse strategies for 3D scene tokenization and generation. Methods like OccWorld~\cite{occworld} employ VQ-VAE-based tokenization paired with autoregressive transformers to predict future scenes. RenderWorld~\cite{renderworld} introduces an AM-VAE that separates empty and non-empty voxels for structured generation. Language-augmented approaches such as OccLLaMA~\cite{occllama} and Occ-LLM~\cite{occ-llm} integrate 3D occupancy tokens with textual scene descriptions to guide generation via large language models. Other works like DriveWorld~\cite{driveworld} and UniWorld~\cite{uniworld} leverage 4D occupancy reconstruction for pretraining. Drive-OccWorld~\cite{drive-occworld} incorporates action conditioning for controllable scene synthesis. To improve efficiency, DFIT-OccWorld~\cite{dfit-occworld} adopts a non-autoregressive pipeline, whereas OccSora~\cite{occsora} and DOME~\cite{dome} utilize 4D tokenizers with diffusion models for long-horizon generation. UniScene~\cite{uniscene} proposes a temporal-aware VAE which largely improve the reconstruction performance. Despite these advances, existing methods struggle to jointly optimize tokenization efficiency and dynamic fidelity.
    \subsection{Tokenizers for Generation}
        Previous studies on image generation~\cite{vq-vae,vq-vae-v2,vqgan,vqgan-lc,DiT} and video generation~\cite{cv-vae,omnitokenizer,cosmos,videogpt,ivideogpt,magvit,hitvideo,vidtok,v-jepa2} demonstrate the critical role of tokenizers in generative modeling. Most video generation methods~\cite{cosmos,magvit} apply 3D operation based tokenization to capture temporal dependencies. Concurrently, iVideoGPT~\cite{ivideogpt} proposes a dual encoder–decoder architecture that independently processes the conditioning frame and the subsequent frame to mitigate temporal redundancy. HiTVideo~\cite{hitvideo} encodes video sequences into hierarchically structured codebooks. In this work, we introduce a novel temporal-quantization method that models intra- and inter-scene information to provide an effective generative representation.

\section{Method}
    \begin{figure*}[t]
    \centering
    \includegraphics[width=0.98\textwidth]{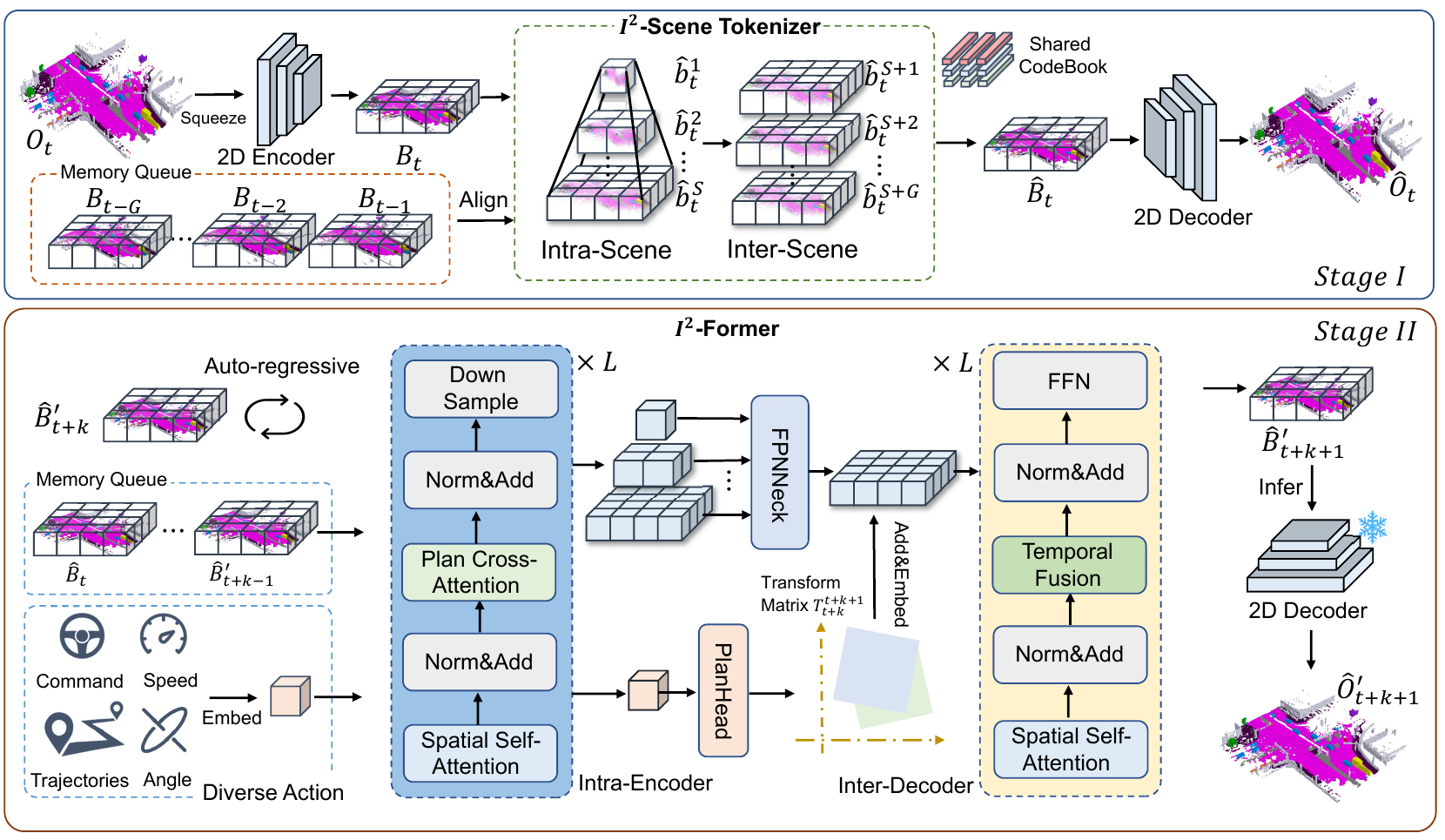}
    \vspace*{-0.1cm}
    \caption{ 
    \textbf{The overall architecture of $I^{2}$-World.} Our framework consists of two core components: (1) the $I^{2}$-Scene Tokenizer, which dynamically converts 3D scene data into compact tokens to enable efficient representation learning; and (2) the $I^{2}$-Former, an autoregressive transformer conditioned on transformation matrices that iteratively leverages the compact tokens for forecasting.
    }
    \label{label:model_framework}
    \vspace{-1em}
\end{figure*}

    \subsection{Overall Architecture}
        As illustrated in \cref{label:model_framework}, our model comprises two major components: the $I^{2}$-Scene Tokenizer and the $I^{2}$-Former. At each timestep $t$, given the current occupancy $O_{t} \in \mathbb{R}^{H \times W \times Z}$ and the historical observation sequence $\{O_{t-g}\}_{g=1}^{G}$ of length $G$, our model first projects $\{O_{t-g+1}\}_{g=1}^{G+1}$ into a latent space via the $I^{2}$-Scene Tokenizer, yielding compressed continuous tokens $\hat{B}_{t} \in \mathbb{R}^{h\times w\times C}$, where $h,w \ll H,W$. The $I^{2}$-Former then autoregressively predicts $K$ latent tokens $\{\hat{B}^{'}_{t+k} \}_{k=1}^{K}$ that are subsequently decoded into the occupancy $\{\hat{O}^{'}_{t+k} \}_{k=1}^{K}$.
        Our model follows a two-stage training procedure: in the first stage, we adopt the VAE~\cite{vq-vae} pipeline to train the $I^{2}$-Scene Tokenizer (detailed in~\cref{sec: scene_tokenizer}); in the second stage, we freeze the $I^{2}$-Scene Tokenizer and learn the dynamic transition for the $I^{2}$-Former (detailed in~\cref{sec: former}). It should be noted that, unlike prior generation methods~\cite{vista, occsora, dome} that rely on trajectory-based guidance, our approach employs a transformation matrix for fine-grained control, enabling higher-fidelity generation through explicit spatial constraints.

    \subsection{\texorpdfstring{$I^{2}$} --Scene Tokenizer}
        \label{sec: scene_tokenizer}
        Traditional scene tokenization frameworks~\cite{vq-vae, cv-vae, occworld, renderocc, dome} follow an encoding-quantization-decoding pipeline. Given an input 3D scene $O_{t} \in \mathbb{R}^{H \times W \times Z}$ at timestep $t$, an encoder first compresses $O_{t}$ into a downsampled feature map $B_{t} \in \mathbb{R}^{h \times w \times C}$, where $h = \frac{H}{d}$, $w = \frac{W}{d}$, and $d$ is the downsampling factor. Subsequently, a quantizer $\mathcal{Q}(\cdot)$ maps each spatial feature vector $B_{t}^{i,j} \in \mathbb{R}^{C}$ to the nearest code in a learnable codebook $\mathcal{C} \in \mathbb{R}^{N \times C}$, where $B_{t}^{i,j}$ denotes the vector at location $(i,j)$ and $\mathcal{C}$ contains $N$ codes of dimension $C$. Each code $\mathbf{c} \in \mathcal{C}$ represents a high-level concept of the scene. The quantization is performed as follows:
        \begin{equation}
            \hat{B}^{i,j}_{t} = \mathcal{Q}(B^{i,j}_{t}, \mathcal{C}) = \underset{\mathbf{c} \in \mathcal{C}}{\arg\min} \, \| B^{i,j}_{t} - \mathbf{c} \|_{2},
        \end{equation}
        where $||\cdot||_{2}$ denotes the $L_2$ norm. The quantized tokens $\hat{B}^{i,j}_{t}$ are aggregated into a token map $\hat{B}_{t}$, which serves as the compact scene representation. Finally, a decoder reconstructs the original scene $\hat{O}_{t}$ from $\hat{B}_{t}$, minimizing information loss during tokenization.

        However, this traditional approach fails to explicitly model temporal dynamics. Instead, temporal information is crudely approximated by stacking tokens across timesteps (e.g., $\hat{B}_{1:t} \in \mathbb{R}^{t\times h\times w\times C}$) and relying on computationally costly spatiotemporal attention mechanisms~\cite{occworld, bevformerv2, dome}. This inefficient strategy increases memory overhead and limits real-time scalability. Moreover, utilizing the discrete tokens to represent 3D space which also hurt the continuous representation making it hard for keeping spatial consistency. To overcome these limitations, we propose the continuous-based $I^{2}$-Scene Tokenizer, which decouples tokenization into two complementary processes: intra-scene tokenization for spatial granularity and inter-scene tokenization for explicit temporal modeling.  

        \paragraph{Intra-Scene Tokenization.} The Intra-Scene Tokenization aims to capture fine-grained spatial context within individual timesteps. Given that 3D occupancy inherently supports multi-scale representation through adjustable voxel resolutions, we adopt a hierarchical multi-scale quantization layer inspired by the residual design of RQ-VAE \cite{rq-vae} and VAR~\cite{var}. Starting with a feature map $B_{t}$, we iteratively quantize it into $S$ multi-scale token maps $\{b^{s}_{t}\}_{s=1}^{S}$. At each scale $s$, the token map $b^{s}_{t}\in \mathbb{R}^{h_{s} \times w_{s} \times C}$ spans a higher-resolution grid than $b^{s-1}_{t}$ and the final token map $b^{s}_{t}$ matches the original feature map’s resolution $h \times w$. Crucially, each token map $b^{s}_{t}$ depends on its preceding map $\{b^{s^{'}}_{t}\}_{s^{'}=1}^{s-1}$ and a single shared codebook $\mathcal{C}$ is applied across all scales. For each scale $s$, the quantization process is defined as follows:
        \begin{equation}
            \begin{split}
                b^{s}_{t} &= f_{intp}(B_{t}, h_{s}, w_{s}),      \\
                \hat{b}^{s_{i,j}}_{t}&=\mathcal{Q}(b^{s_{i,j}}_{t}, \mathcal{C}), \\
                B_{t}&=B_{t}-f_{intp}(\hat{b}^{s}_{t}, h, w), \\
                \hat{B}_{t}&=\hat{B}_{t}+\phi_{s}(\hat{b}^{s}_{t}).
            \end{split} 
        \end{equation}
        
        The function $f_{intp}(B,h,w)$ resizes feature map $B$ to the target spatial resolution $h\times w$, $\hat{b}^{s_{i,j}}_{t}$ represents the vector at position $(i,j)$ of $\hat{b}^{s}_{t}$ and $\hat{B}_{t}$ denotes the output token map which is initialized to zero at the start scale, and $\phi_{s}$ denotes a learnable convolution layer that mitigates information loss during resolution scaling. After encoding the static spatial information of $B_{t}$ into $\hat{B}_{t}$, we apply Inter-Scene Tokenization to explicitly model temporal dynamics across timesteps. 

        \paragraph{Inter-Scene Tokenization.} To model temporal dynamics, we maintain a memory queue storing $G$ historical feature maps $\{B_{t-g}\}_{g=1}^{G}$. First, at each timestep $t$, we align historical features $B_{t-g}$ to the current scene using ego-pose transformation matrix $T_{t-g}^{t}$ :
        \begin{equation}
            B_{t-g}^{'} = T_{t-g}^{t} B_{t-g}.
        \end{equation}
        
        Next, the residual features of $B_{t}$ from Intra-Scene Tokenization are combined with aligned historical maps $\{B_{t-g}^{'}\}_{g=1}^{G}$ to iteratively generate $G$ temporal token maps $\{b^{S+g}_{t}\}_{g=1}^{G}$. Each token map $b^{S+g}_{t} \in \mathbb{R}^{h\times w \times C}$ depends on its predecessors $\{b^{S+g^{'}}_{t}\}_{g^{'}=1}^{g-1}$, and the same codebook $\mathcal{C}$ from Intra-Scene Tokenization ensures spatial-temporal continuity. The quantization at step $g$ is:
        \begin{equation}
            \begin{split}
                b_{t}^{S+g} &= B_{t}+B_{t-g}^{'}, \\
                \hat{b}_{t}^{(S+g)_{i,j}}&= \mathcal{Q}(b_{t}^{(S+g)_{i,j}}, \mathcal{C}),\\
                \hat{B}_{t} &= \hat{B}_{t} + \psi_{g}(\hat{b}_{t}^{S+g}), \\
                B_{t} &= B_{t} - \hat{b}_{t}^{S+g}.\\
            \end{split}
        \end{equation}
        
        $\psi_{g}$ is a learnable convolution layer mitigating information loss during temporal aggregation and $\hat{b}^{(S+g)_{i,j}}_{t}$ represents the vector at position $(i,j)$ of $\hat{b}_{t}^{S+g}$. The output $\hat{B}_{t}$ retains the compression efficiency of traditional single-stage tokenization while preserving temporal dynamics critical for scene generation. Compared to vanilla tokenizer architectures, $I^{2}$-Scene Tokenizer achieves significant performance gains with only $G+S$ lightweight convolution layers. This balance between minimal architectural overhead and improved reconstruction fidelity underscores the effectiveness of decoupled intra- and inter-scene tokenization.  

        \paragraph{Encoder \& Decoder.} To enhance computational efficiency, we adopt an encoder–decoder architecture similar to that of OccWorld~\cite{occworld}. The encoder processes the occupancy $O_{t}$ into a bird’s-eye-view representation and employs a ResNet augmented with attention mechanisms~\cite{ldm, attention_is_all_you_need} for both encoder and decoder. At the decoder terminus, the channel dimension is partitioned to restore the height information of the reconstruction occupancy data $\hat{O}_{t}$. 

        \paragraph{Training Loss.} To supervise 3D occupancy reconstruction and codebook learning, we optimize a composite loss comprising a weighted focal loss \cite{fb-occ}, Lovasz loss~\cite{lovasz}, and vector quantization loss \cite{vq-vae}:
        \begin{equation}
            \begin{split}
                \mathcal{L}_{token} &= \mathcal{L}_{focal}(O_{t}, \hat{O}_{t})+\mathcal{L}_{lov}(O_{t}, \hat{O}_{t}) + \mathcal{L}_{vq}, \\
                \mathcal{L}_{vq} &= \sum_{s=1}^{S} \sum_{i,j} ||sg(b_{t}^{s_{i,j}})-\hat{b}_{t}^{s_{i,j}}||_{2}^{2} + \\
                \beta &||b_{t}^{s_{i,j}}-sg(\hat{b}_{t}^{s_{i,j}})||_{2}^{2},
            \end{split}
        \end{equation}
        where $sg(\cdot)$ denotes the stop-gradient operation and $\beta$ is a hyperparameter defaulting to 1. It should be noted that we supervise only the Intra-Scene Tokenization for codebook learning for stable learning. Upon completing tokenizer training, the compact latent representation $\hat{B}_{t}$ is fed into the proposed $I^{2}$-Former to predict future scene states.

    \subsection{\texorpdfstring{$I^{2}$} --Former}
        \label{sec: former}
        Unlike prior decoder-only methods~\cite{occ-llm, occllama, occworld}, $I^{2}$-Former employs an encoder-decoder architecture. For each generation timestamp $t+k$, we first encode diverse historical ego-motion information (e.g., speed, trajectories) into a plan embedding using a multi-layer perceptron (MLP). The plan embedding and the generated latent token $\hat{B}_{t+k}^{'}$ are jointly processed by the Intra-Encoder, which hierarchically aggregates spatial context and fuses plan–scene interactions via cross-attention mechanisms~\cite{attention_is_all_you_need}. The fused representation of plan embedding from the final layer $L$ subsequently guides estimation of the transformation matrix $T_{t+k}^{t+k+1} \in \mathbb{R}^{4\times4}$ that maps the scene state from timestep from $t+k$ to $t+k+1$. Finally, the Inter-Decoder utilizes $T_{t+k}^{t+k+1}$ as a spatiotemporal condition to autoregressively predict the subsequent latent token $\hat{B}_{t+k+1}^{'}$, thereby iteratively evolving the scene state.

        \paragraph{Intra-Encoder.} At each layer $l$, the Intra-Encoder employs spatial self-attention (SSA) \cite{deformable-detr, bevformer} to model adaptive interactions between spatial tokens in $\hat{B}_{t+k}^{'}$. For a query feature $Q_{x,y}$ at position $p=(x,y)$, the attention mechanism dynamically aggregates contextual information from a sparse set of relevant tokens $V\in \hat{B}_{t+k}^{'}$, weighted by their geometric and semantic relevance. Formally, this is expressed as:
        \begin{equation}
            SSA(Q_{x,y}, \hat{B}_{t+k}^{'}) = \sum_{_{V \in \hat{B}_{t+k}^{'}}} F (Q_{x,y}, p, V),
        \end{equation}
        where $F$ denotes the deformable attention operation \cite{deformable-detr}, which learns to sample and attend to critical regions of the token map $\hat{B}_{t+k}^{'}$. This mechanism allows tokens to flexibly interact based on their spatial relationships and encoded features, prioritizing regions with high structural or semantic relevance.

        To integrate scene information with the plan embedding, we employ a multi-head attention mechanism \cite{attention_is_all_you_need}, enabling the plan embedding to interact with scene tokens across both spatial and semantic dimensions. At the end of each hierarchical layer $l$, $\hat{B}_{t+k}^{'}$ is downsampled by a factor of two, yielding progressively coarser token maps. This multi-scale hierarchy allows the plan embedding to adaptively fuse coarse-to-fine spatial context. By default, we use $L=3$ layers, balancing computational efficiency and granularity preservation.  

        The Intra-Encoder outputs multi-scale token maps and a refined plan embedding. To fuse multi-scale spatial context, we apply a lightweight FPN~\cite{fpn}, aggregating hierarchical features into a unified representation feature. Simultaneously, the plan embedding is used to regress a transformation matrix $T_{t+k}^{t+k+1}$, which encodes the spatiotemporal shift between consecutive timesteps. The matrix $T_{t+k}^{t+k+1}$ serves as a conditioning signal for autoregressive token prediction. We project $T_{t+k}^{t+k+1}$ into the latent space via a MLP and add it to the unified representation feature, ensuring the Inter-Decoder leverages both scene geometry and planned dynamics to forecast $\hat{B}_{t+k+1}^{'}$.

        \paragraph{Inter-Decoder.} The Inter-Decoder follows the conventional transformer architecture~\cite{attention_is_all_you_need} but introduces two tailored designs: SSA and Lightweight Temporal Fusion. To enforce temporal consistency during autoregressive generation, the Inter-Decoder maintains a memory queue $\{ \hat{B}_{t}, \hat{B}_{t+1}^{'} \cdots \hat{B}_{t+k-1}^{'} \}$ of historical scene tokens. At each layer $l$, features are refined by SSA which adaptively aggregating contextual cues from the conditioned token map, ensuring spatial consistency within the scene. Subsequently, since temporal dynamics are already embedded in the scene tokens, we concatenate all historical tokens along the channel dimension and apply a single-layer MLP to fuse temporal context. This minimalist design ensures token consistency without introducing compute-heavy cross-attention or recurrent modules. Finally, the output of Inter-Decoder, $\hat{B}_{t+k+1}^{'}$ is appended to the memory queue, propagating temporal context to subsequent autoregressive steps.

        \paragraph{Training Loss.} To supervise the autoregressive generation, we utilize the mean squared error (MSE) to measure the feature-level similarity between the predicted token map $\hat{B}_{t+k+1}^{'}$ and the tokenizer-generated token map $\hat{B}_{t+k+1}$. We assign different weights along the timestamp to avoid the cumulative error affecting the training process:
        \begin{equation}
            \mathcal{L}_{gen} = \sum_{k=1}^{K}w_{k} \mathcal{L}_{mse} (\hat{B}_{t+k}^{'}, \hat{B}_{t+k}),
        \end{equation}
        where $w_{k}$ denotes hyperparameters for different frames. For the supervision of the conditional transformation matrix, we decompose the transformation matrix into its translation and rotation components. We apply the L2 loss to supervise the translation and the cosine loss to supervise the quaternion representation of rotation.

        \paragraph{Controllable Generation.} Our method provides two controllable strategies for 4D occupancy generation. First, users can modulate diverse ego-vehicle actions (e.g., command, speed profiles) to exert high-level trajectory control. Second, the transformation matrix enables direct manipulation of spatial-temporal transitions, allowing fine-grained guidance of scenario-specific generation.

\section{Experiments}
    \begin{table*}[!t]
    \setlength{\tabcolsep}{0.0140\linewidth}
    \centering
    \scalebox{0.8}{
    \begin{tabular}{l|c|cccc>{\columncolor{Gray}}c|cccc>{\columncolor{Gray}}c|cc}
        \toprule
        \multirow{2}{*}{Method} & \multirow{2}{*}{Input} & \multicolumn{5}{c|}{mIoU (\%) $\uparrow$} &  \multicolumn{5}{c|}{IoU (\%) $\uparrow$} & \\
        & & Recon. & 1s & 2s & 3s &  Avg. & Recon. &1s & 2s & 3s & Avg. & \multirow{-2}*{FPS} \multirow{-2}*{$\uparrow$} \\
        \midrule
        OccWorld-O~\cite{occworld}     & Occ               & 66.38 & 25.78 & 15.14 & 10.51 & 17.14 & 62.29 & 34.63 & 25.07 & 20.18 & 26.63 & 18.00 \\
        OccLLaMA-O~\cite{occllama}     & Occ \& Text       & 75.20 & 25.05 & 19.49 & 15.26 & 19.93 & 63.79 & 34.56 & 28.53 & 24.41 & 29.17 & -   \\
        Occ-LLM~\cite{occ-llm} & Occ \& Text & - & 24.02 & 21.65 & 17.29 & 20.99 & -& 36.65 &32.14 & 28.77 & 32.52 & - \\
        DFIT-OccWorld~\cite{dfit-occworld}  & Occ \& Camera & -     & 31.68 & 21.29 & 15.19 & 22.71 & -     & 40.28 & 31.24 & 25.29 & 32.27 & -   \\
        DOME~\cite{dome}             & Occ               & 83.08 & 35.11 & 25.89 & 20.29 & 27.10 & 77.25 & 43.99 & 35.36 & 29.74 & 36.36 & 6.54   \\
        UniScene~\cite{uniscene}     & Occ               & \textbf{92.10}  & 35.37 & 29.59 & 25.08 & 31.76 & \textbf{87.00}  & 38.34 & 32.70 & 29.09 & 34.84 & -    \\
        \rowcolor{aliceblue}
        $I^{2}$-World-O (Ours)         & Occ     & 81.22 & \textbf{47.62} & \textbf{38.58} & \textbf{32.98} & \textbf{39.73} & 68.30 & \textbf{54.29} & \textbf{49.43} & \textbf{45.69} & \textbf{49.80} & \textbf{37.04} & \\   
        \midrule
        OccWorld-STC                 & Camera   & 23.63     & 10.97 & 15.16 & 10.19 & 7.53 & 31.20 & 20.68 & 24.81 & 20.09 & 17.14 & 3.77 \\
        DOME-STC                     & Camera   & \textbf{25.30} & 17.79 & 14.23 & 11.58 & 14.53 & \textbf{33.15} & 26.39 & 23.20 & 20.42 & 23.33 & 2.75 \\
        \rowcolor{aliceblue}
        $I^{2}$-World-STC (Ours)     & Camera   & 25.13 & \textbf{21.67} & \textbf{18.78} & \textbf{16.47} & \textbf{18.97} & 32.66 & \textbf{30.55} & \textbf{28.76} & \textbf{26.99} & \textbf{28.77} & \textbf{4.21} \\
        \bottomrule
    \end{tabular}%
    }
    \caption{\textbf{4D occupancy forecasting performance on Occ3d-nus validation dataset.} Avg. denotes average performance of that in 1s, 2s, and 3s. The symbol ``-''
 indicates that the method either does not report results or is not open-sourced.}
    \label{tab:mainresults_4dforecast}
\end{table*}
    \begin{figure*}[t]
    \centering
    \includegraphics[width=0.96\textwidth]{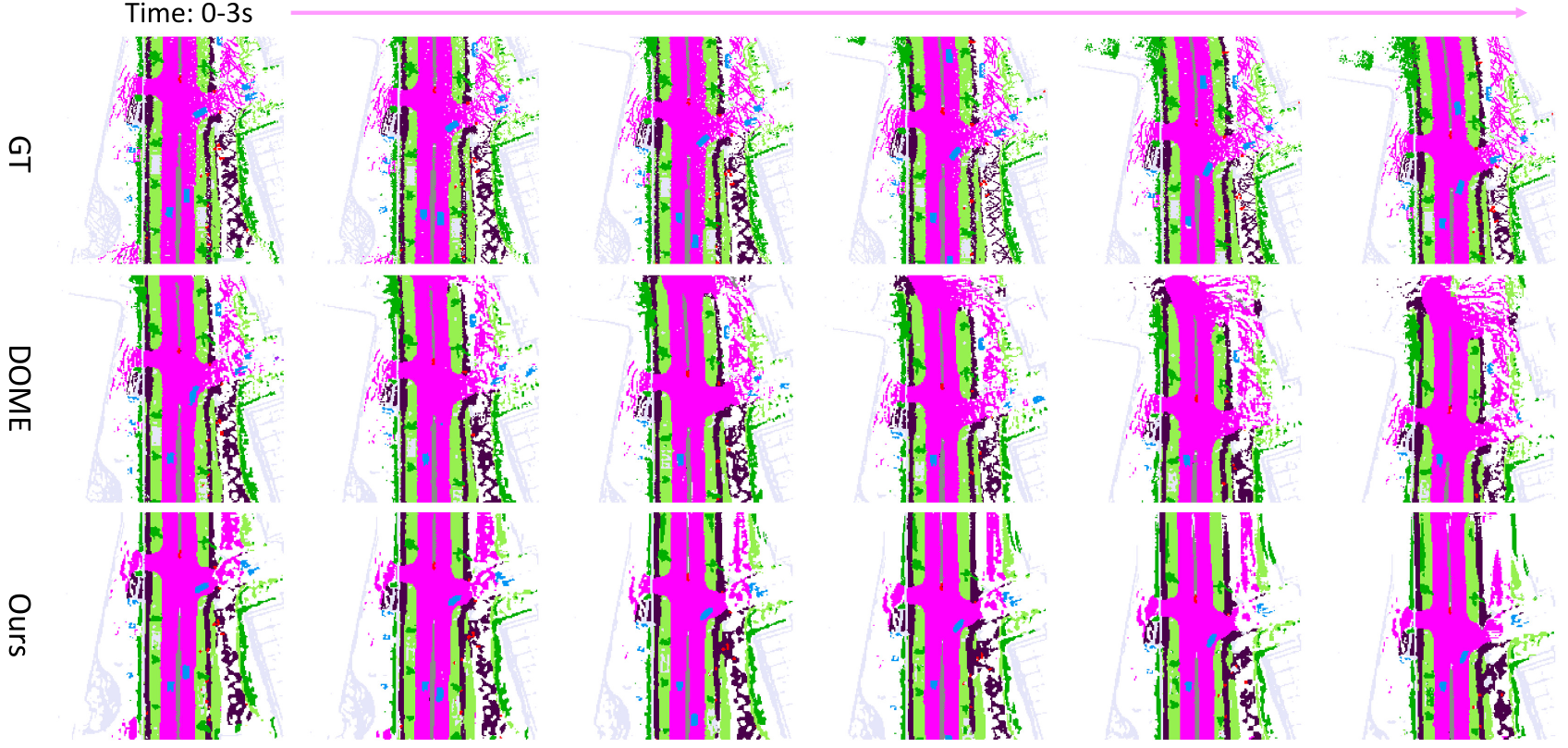}
    \vspace*{-0.1cm}
    \caption{ 
    \textbf{Qualitative results of 4D occupancy forecasting on Occ3d-nus validation set.}
    }
    \label{figure:visualization_4d_occupancy_forecasting}
    \vspace{-1em}
\end{figure*}

    \subsection{Experimental Setup}
        \paragraph{Dataset.} To evaluate the forecasting capability of our method, we utilize the Occ3D-nus benchmark~\cite{occ3d}, a 4D occupancy forecasting dataset derived from the nuScenes autonomous driving dataset~\cite{nuscenes}. Occ3D-nus contains 600 training scenes and 150 validation scenes. The spatial coverage of each sample spans $[-40,\text{m}, 40,\text{m}]$ along the x- and y-axes and $[-1,\text{m}, 5.4,\text{m}]$ along the z-axis, with a uniform voxel resolution of 0.4 m in all dimensions and a sampling rate of 2 Hz. To evaluate the generalization of our model, we utilize Occ3D-Waymo~\cite{waymo, occ3d}, comprising 202 validation scenes with the same spatial range and voxel resolution but a sampling rate of 10 Hz.

        \paragraph{Implementation Details.} Aligning with existing works~\cite{occworld, dfit-occworld, occ-llm, dome}, we utilize 2-second historical data to forecast 3-second future occupancy. Our proposed $I^{2}$ Scene Tokenizer leverages a shared codebook containing 512 entries and 128-dimensional latent features to encode compact spatio-temporal representations. Training employs the AdamW optimizer~\cite{adamw} with a base learning rate of $1\times10^{-3}$, a global batch size of 128, and distinct schedules for each component: the $I^{2}$ Scene Tokenizer is trained for 24 epochs, while the $I^{2}$Former undergoes 48 epochs of training. For evaluation, we employ the ground-truth transformation matrix to guide generation and adopt mIoU and IoU as evaluation metrics for the 4D occupancy forecasting task. The FPS calculation method follows OccWorld~\cite{occworld} and is measured using an RTX 4090 GPU.

    \subsection{Main Results}

         \paragraph{4D Occupancy Forecasting.} Following established evaluation protocols~\cite{occworld, dfit-occworld, dome}, we assess our $I^{2}$-World framework in two configurations: (1) $I^{2}$-World-O using ground-truth 3D occupancy inputs, and (2) $I^{2}$-World-STC utilizing predictions from STCOcc~\cite{stcocc}. As demonstrated in \cref{tab:mainresults_4dforecast}, $I^{2}$-World-O achieves substantial gains of 25.1$\%$ mIoU (39.73 vs. 31.75) and 36.9$\%$ IoU (49.80 vs. 36.36) over current state-of-the-art approach. Notably, whereas existing solutions rely on resource-intensive architectures such as large language models or diffusion models, the proposed system maintains exceptional efficiency, requiring only 2.9 GB of training memory and achieving real-time inference at 37.04 FPS. The end-to-end $I^{2}$-World-STC variant shows particularly promising results, outperforming prior methods by 50.9$\%$ (18.97 vs. 12.57) in mIoU and 40.9$\%$ (28.77 vs. 20.41) in IoU metrics.

        \begin{table}[tp]
    \centering
    \scalebox{0.77}
    {
        \begin{tabular}{c|c|cc|cc}
        \toprule
         \multirow{2}{*}{Method} & \multirow{2}{*}{Rate} & \multicolumn{2}{c|}{Reconstruction} & \multicolumn{2}{c|}{Forecasting}\\
                                 &                             & mIoU($\%$)$\uparrow$ & IoU($\%$)$\uparrow$ & mIoU($\%$)$\uparrow$ & IoU($\%$)$\uparrow$ \\
        \midrule
        Copy-Paste          &    10Hz      & \textbf{76.27}          & \textbf{74.62}           &  28.34  & 40.09 \\
        \rowcolor{aliceblue}
        $I^{2}$-World       &    10Hz      & \textbf{76.27} & \textbf{74.62}  & \textbf{43.73}    &  \textbf{60.97}  \\
        \midrule
        Copy-Paste          &    2Hz      & \textbf{76.27} & \textbf{74.62}           &  17.17  & 28.21 \\
        \rowcolor{aliceblue}
        $I^{2}$-World       &    2Hz     & \textbf{76.27} & \textbf{74.62}  & \textbf{36.38}    &  \textbf{52.36}  \\
        \bottomrule
        \end{tabular}
    }
    \caption{\textbf{Zero-shot performance in Occ3D-Waymo dataset.}}
    \label{tab:zero-shot on waymo}
\end{table}
        \paragraph{Generalization Ability.} To evaluate the generalization of our method, we conduct zero-shot 4D occupancy forecasting on the Occ3D-Waymo dataset~\cite{occ3d, waymo}. The copy-paste baseline leverages current-frame reconstruction for forecasting. As shown in~\cref{tab:zero-shot on waymo}, our method is evaluated under multiple sampling rates. The result demonstrates excellent generalization on the Waymo dataset~\cite{waymo}, suggesting its potential as an auto-labeling method.

    \subsection{Ablation Study}
        \begin{table}[!tp]
    \setlength{\tabcolsep}{0.016\linewidth}
    \centering
    \scriptsize
  \scalebox{1.0}{
  \begin{tabular}{ccc|cc|cc}
    \toprule
    \multirow{2}{*}{Baseline}    & \multicolumn{2}{c|}{Inter-Scene} & \multicolumn{2}{c|}{Intra-Scene} & \multirow{2}{*}{mIoU($\%$)$\uparrow$} & \multirow{2}{*}{IoU($\%$)$\uparrow$} \\
                &  w/o Align      & w Align         &  S-Scale          &   M-Scale  &                      &                      \\
    \midrule
    \ding{51}   &               &                   &  \ding{51}        &              &          66.52      &  61.07               \\
    \ding{51}   &   \ding{51}   &                   &  \ding{51}        &              &          70.37      &  62.18               \\
    \ding{51}   &               &    \ding{51}      &  \ding{51}        &              &          77.12      &  64.20               \\
    \rowcolor{aliceblue}
    \ding{51}   &              &    \ding{51}       &                   &    \ding{51} & \textbf{81.22}      &  \textbf{68.30}       \\
    \bottomrule
  \end{tabular}
  }
  \caption{\textbf{Ablation on $I^{2}$Scene Tokenizer.} ``w Align'' and ``w/o Align'' indicate whether a transformation matrix is used to align historical feature maps. The mIoU and IoU metrics evaluate the reconstruction accuracy of the scene.}
  \label{table:ablation-tokenizer}
  \vspace{-1.5em}
\end{table}
        \paragraph{The Effectiveness of $I^{2}$ Scene Tokenizer.} 
        As shown in \cref{table:ablation-tokenizer}, we conduct an ablation study on the $I^{2}$ Scene Tokenizer to evaluate the contribution of each design component. Our baseline adopts a single-scale Scene Tokenizer from OccWorld~\cite{occworld}, augmented with a flipping BEV plane to enhance tokenizer generalization. The Inter-Scene tokenization module improves temporal modeling, yielding gains of 5.7$\%$ mIoU and 1.8$\%$ IoU, while the alignment strategy further enhances temporal information modeling, yielding gains of 15.9$\%$ mIoU and 5.1$\%$ IoU. Notably, Inter-Scene tokenization operates as a plug-and-play module compatible with other tokenizers. By combining Inter-Scene tokenization with the multi-scale Intra-Scene tokenizer, we achieve optimal performance. The results demonstrate that decoupled spatial-temporal tokenization effectively addresses distinct aspects of 3D scene understanding while enabling efficient, synergistic holistic modeling.

        \begin{table}[tp]
    \resizebox{\linewidth}{!}{

        \begin{tabular}{ccc|cc|ccc}
        \toprule
        \multicolumn{3}{c|}{Inter-Decoder} & \multicolumn{2}{c|}{Intra-Encoder} & \multicolumn{3}{c}{Metric} \\
        \midrule
        
        Trans & Rotate & TF & With & MS & mIoU($\%$)$\uparrow$ & IoU($\%$)$\uparrow$ & Mem(G)$\downarrow$ \\
        \midrule
                  &            &             &          &  & 17.12   & 27.75      & \textbf{1.81}       \\
        \ding{51} &            &             &          &  & 28.74   & 36.44      &  \textbf{1.81}       \\
                  &  \ding{51} &             &          &  & 20.34   & 29.23      &  \textbf{1.81}        \\
        \ding{51} &  \ding{51} &             &          &  & 34.25    & 42.71     &  \textbf{1.81}        \\
        \ding{51} &  \ding{51} & \ding{51}   &          &  & 35.21   & 43.32      &  2.03          \\
        \ding{51} & \ding{51}  & \ding{51}   & \ding{51}&  & 36.73   & 46.84      &  3.74        \\
        \rowcolor{aliceblue}
        \ding{51} & \ding{51}  & \ding{51}   & \ding{51}& \ding{51}& \textbf{39.73}   & \textbf{49.80} & 2.92        \\
        \bottomrule
        \end{tabular}

    }
    \caption{\textbf{Ablation study on the each component of $I^{2}$Former.} We decompose the transformation matrix into translation and rotation components, where TF denotes the temporal fusion module, ``With'' indicates the use of the Intra-Encoder, and MS signifies the adoption of a multi-scale strategy within the Intra-Encoder. The mIoU and IoU metrics evaluate the 4D forecasting of the scene.}
    \label{tab:ablation_former}
    \vspace{-1em}
\end{table}
        \paragraph{The Effectiveness of $I^{2}$ Former.} As shown in \cref{tab:ablation_former}, we analyze the contributions of individual components in the Inter-Decoder and Intra-Encoder. First, we evaluate a baseline configuration where the vanilla Inter-Decoder predicts future occupancy without conditioning, achieving 17.1$\%$ mIoU and 27.2$\%$ IoU. This result highlights the latent token's inherent capacity to encode dynamic scene information, albeit with limited predictive accuracy. Next, we introduce translation as a conditioning signal while keeping rotation fixed to the start frame. This modification achieves a 67.8$\%$ mIoU and 31.4$\%$ IoU, representing a performance gain of 50.7$\%$ mIoU and 4.2$\%$ IoU over the baseline. In contrast, introducing rotation as the sole conditioning signal yields only 18.7$\%$ mIoU and 5.4$\%$ IoU, demonstrating that rotation alone contributes minimally to performance. Finally, combining both translation and rotation conditioning leads to a significant improvement, achieving 85.3$\%$ mIoU and 38.9$\%$ IoU, which underscores the complementary role of these spatial transformations in modeling scene dynamics. The Intra-Encoder, lightweight temporal fusion, and multi-scale strategy incur only a limited increase in GPU memory while delivering gains of 16.0$\%$ mIoU and 16.6$\%$ IoU, demonstrating the efficiency and effectiveness of our design.

    \begin{figure*}[t]
    \centering
    \includegraphics[width=0.98\textwidth]{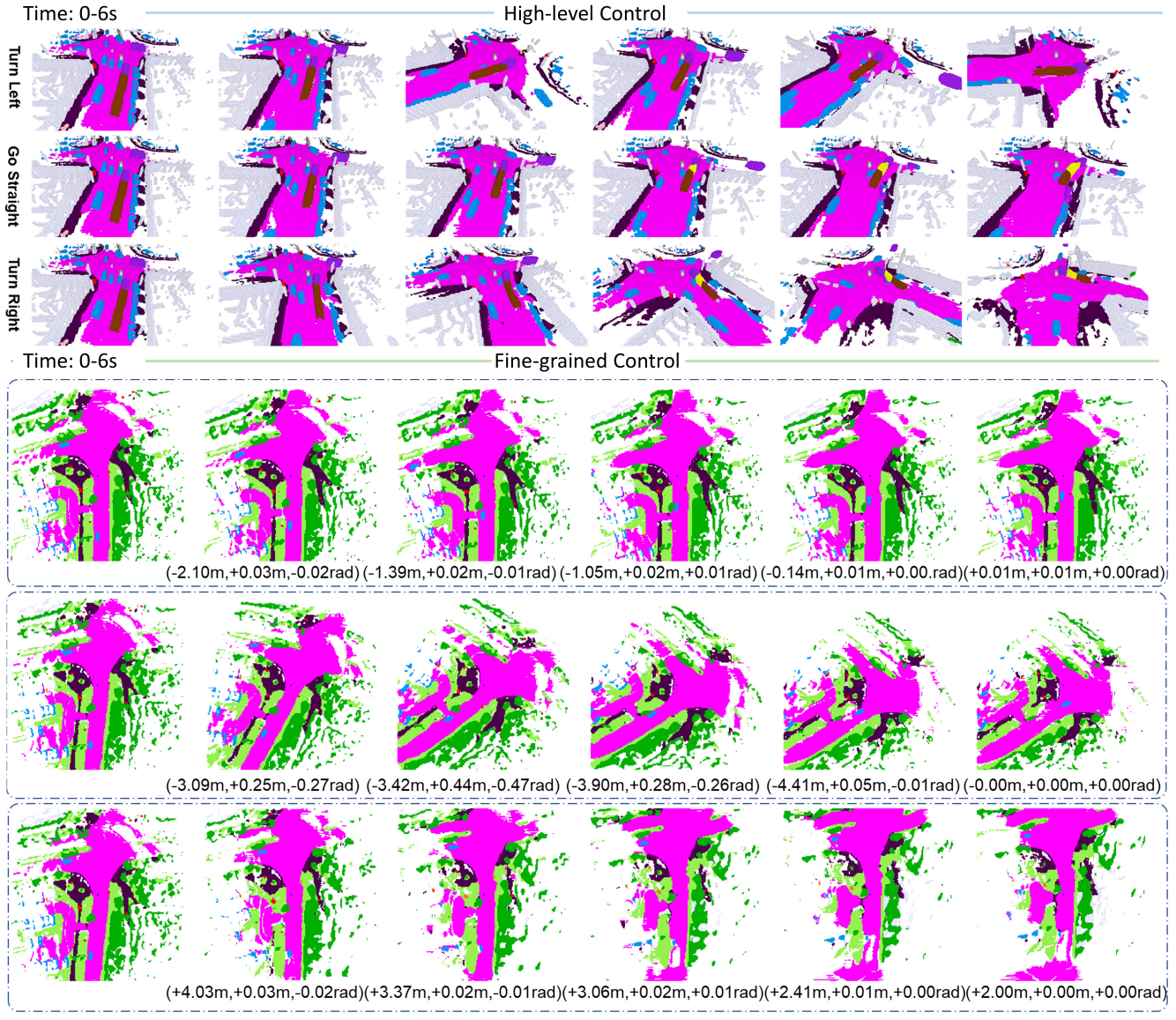}
    \vspace*{-0.1cm}
    \caption{ 
    \textbf{Demonstration of the controllability of our methods.} The first column in each case represents the conditioned frame.
    }
    \label{figure:visualization_control_generation}
    \vspace{-1em}
\end{figure*}

    \subsection{Visualizations}
        \paragraph{4D Occupancy Forecasting.} In \cref{figure:visualization_4d_occupancy_forecasting}, we visualize 4D occupancy forecasting results on the Occ3D-nuScenes validation set. Compared to DOME \cite{dome}, our method maintains better spatial coherence and more accurately models the movement of moving objects in the 3D scene, thereby demonstrating our method's superior temporal dynamic modeling capabilities.

        \paragraph{Controllable Generation.} \cref{figure:visualization_control_generation} demonstrates our framework's controllability through both high-level commands and fine-grained transformations. For high-level control, distinct commands (e.g., 'turn left/right') generate 6-second scenarios where the ego vehicle's actions dynamically affect surrounding agents. For instance, a 'turn right' command leads to a truck collision, demonstrating our model's capacity to simulate complex interactions and corner cases. 
        
        Fine-grained control employs direct transformation matrix manipulation, enabling precise scene generation (meter/radian-level accuracy). The bottom row presents a failure case: a reversing scenario represented by a transformation matrix (unseen during training) results in unrealistic static agent behavior. We believe that this issue can be mitigated by diversifying the distribution of transformation matrices in the dataset.
        
\section{Conclusion}

    We present a novel 3D scene tokenization framework for 3D scene generation. Our proposed $I^{2}$-Scene Tokenizer employs multi-scale quantization for intra-scene tokenization and temporal quantization for inter-scene tokenization. The proposed $I^{2}$-Former framework introduces transformation matrix-guided generation, enabling fine-grained control while maintaining generation fidelity. Experiments validate our framework's effectiveness and demonstrate its potential as an automated scene labeling solution.

\section*{Acknowledgments}
    This research was supported by the National Natural Science Foundation of China (No.62088102, No. U23B2060), and the Youth Innovation Team of Shaanxi Universities.

{\small
\bibliographystyle{ieeenat_fullname}
\bibliography{11_references}
}


\end{document}